\documentclass[10pt,journal,compsoc]{IEEEtran}
\usepackage{algorithmic}
\usepackage[ruled,vlined,algo2e]{algorithm2e}
\usepackage{amsmath}
\usepackage{amssymb}
\usepackage{caption}
\usepackage{comment}
\usepackage{fixltx2e}
\usepackage[none]{hyphenat}
\usepackage{textcomp}
\usepackage[normalem]{ulem}

\def\B#1{\boldsymbol #1}
\def\C#1{\mathcal #1}

\usepackage{pgfplots}
\usepgfplotslibrary{polar}
\usepgflibrary{shapes.geometric}
\usetikzlibrary{calc}

\usepackage{array}
\newcolumntype{$}{>{\global\let\currentrowstyle\relax}}
\newcolumntype{^}{>{\currentrowstyle}}

\ifCLASSOPTIONcompsoc
  \usepackage[nocompress]{cite}
\else
  \usepackage{cite}
\fi

\begin{document}



\title{SL$^2$MF: Predicting Synthetic Lethality in Human Cancers via Logistic Matrix Factorization}

\author{
    Yong~Liu,
    Min~Wu$^{*}$,
    Chenghao~Liu,
    Xiao-Li~Li,
    and ~Jie~Zheng$^{*}$

    \IEEEcompsocitemizethanks{
        \IEEEcompsocthanksitem Yong Liu is currently with Joint NTU-UBC Research Centre of Excellence in Active Living for the Elderly (LILY), and Alibaba-NTU Singapore Joint Research Institute, Nanyang Technological University, Singapore. Email: liuysc@acm.org.
        \IEEEcompsocthanksitem Jie Zheng is currently with School of Information Science and Technology, ShanghaiTech University, Shanghai 201210, China. Part of this work was performed while he was an Assistant Professor at School of Computer Science and Engineering, Nanyang Technological University, Singapore 639798. Email: zhengjie@shanghaitech.edu.cn.
        \IEEEcompsocthanksitem Min Wu and Xiaoli Li are currently with the Institute for Infocomm Research (I$^{2}$R), A*Star, Singapore 138632. Email: \{wumin, xlli\}@i2r.a-star.edu.sg.
        \IEEEcompsocthanksitem Chenghao Liu is currently with School of Information Systems, Singapore Management University, Singapore. Email: twinsken@gmail.com.
        \IEEEcompsocthanksitem $^{*}$ Corresponding authors

    }

    \thanks{Manuscript received xxx, 2018; revised xxx, 2018.}
}

\IEEEtitleabstractindextext{%
\begin{abstract}
Synthetic lethality (SL) is a promising concept for novel discovery of anti-cancer drug targets. However, wet-lab experiments for detecting SLs are faced with various challenges, such as high cost, low consistency across platforms or cell lines. Therefore, computational prediction methods are needed to address these issues. This paper proposes a novel SL prediction method, named \textsf{SL$^2$MF}, which employs logistic matrix factorization to learn latent representations of genes from the observed SL data. The probability that two genes are likely to form SL is modeled by the linear combination of gene latent vectors. As known SL pairs are more trustworthy than unknown pairs, we design importance weighting schemes to assign higher importance weights for known SL pairs and lower importance weights for unknown pairs in \textsf{SL$^2$MF}. Moreover, we also incorporate biological knowledge about genes from protein-protein interaction (PPI) data and Gene Ontology (GO). In particular, we calculate the similarity between genes based on their GO annotations and topological properties in the PPI network. Extensive experiments on the SL interaction data from SynLethDB database have been conducted to demonstrate the effectiveness of \textsf{SL$^2$MF}.
\end{abstract}

}

\maketitle
\IEEEdisplaynontitleabstractindextext
\IEEEpeerreviewmaketitle

\section{Introduction}


A complex disease like cancer is unlikely caused by the defect of only one gene. The understanding of genetic interactions, therefore, is becoming more important in cancer biology and medicine \cite{ashworth2011genetic}.  A prominent type of genetic interaction, called synthetic lethality, has drawn much attention in the field of cancer therapeutics \cite{hartwell1997integrating,mclornan2014applying}. A pair of genes is called synthetic lethality (SL) if the defect of a single gene will not affect the cell viability, whereas the defects of both genes will cause cell death or significant impairment of cell fitness. Therefore, targeting  a nonessential SL partner gene of a cancer-specific mutated gene would selectively kill (or prohibit the proliferation of) the cancer cells but spare normal cells. With the availability of ``omics" technologies and high-throughput cancer genomics data, the SLs in the human genome promise to be a gold mine for novel discovery of anti-cancer drug targets. Indeed, both wet-lab screening and computational data mining of SLs in the genomes of human as well as model animal species (\emph{e.g.} yeast) are under intensive research.


High-throughput wet-lab screenings have been conducted to search for SLs genome-wide, using the following technologies. First, chemical libraries are used to identify inhibitors of gene or metabolic activities that can kill cancer cells selectively \cite{simons2001establishment}. Secondly, pooled RNAi screens (using siRNA or shRNA libraries), which target the gene expression at the mRNA level, have been widely adopted for SL detection \cite{turner2008synthetic, luo2009genome}. Considering the complementary strengths of the chemical and RNAi screening technologies, the two types of technologies are sometimes integrated into the more comprehensive approach of chemical-genetic screening \cite{martins2015linking}. Thirdly, the emerging CRISPR-based genome editing technology has been recently employed to screen for essential genes and SLs \cite{du2017genetic, han2017synergistic}. It is expected that this new technology can increase the accuracy and power of screening than the aforementioned technologies. However, the wet-lab screenings for finding SLs are still faced with different challenges, \emph{e.g.}, high cost, off-target effects, lack of consistency across different platforms or cell lines, and unclear mechanisms. Therefore, computational methods for predicting SLs would be useful complements to the wet-lab screenings.


In the literature, various computational methods have been proposed for SL prediction in recent decade \cite{boucher2013genetic, zhan2016towards}. These methods can be classified into the following three categories. The methods in the first category are \emph{in silico} knockouts in metabolic networks. As genome-wide metabolic networks for human and model species are available, single- and double-knockout of genes can be simulated in these networks. By running flux-balance analysis (FBA), the phenotypic cellular effects of these knockouts can be estimated. Based on such metabolic modeling, researchers have carried out predictions for essential and SL genes in yeast \cite{deutscher2008can}, \emph{C.elegans} \cite{suthers2009genome}, human cancer cell lines \cite{folger2011predicting}, as well as other species \cite{pratapa2015fast}. The second category refers to knowledge-based methods \cite{jerby2014predicting,srihari2015inferring,sinha2017systematic}, which predict SLs based on the knowledge or hypotheses about SL interactions, \emph{e.g.}, SL genes tend to be co-expressed, and SL genes are likely to have similar topological properties in biological networks. For example, network properties, \emph{e.g.}, graph centrality \cite{kranthi2013identification}, network flow \cite{zhang2015predicting}, and connectivity homology \cite{jacunski2015connectivity}, are widely explored for SL prediction. Based on the hypotheses that SL genes are often co-expressed and seldom co-mutated, DAISY~\cite{jerby2014predicting} applied three independent procedures for predicting SLs from SCNA (somatic copy number alternation), shRNA and gene expression profiles. Similar to DAISY, MiSL~\cite{sinha2017systematic} also analysed mutation, copy number and gene expression for SL prediction. The third category includes supervised machine learning methods that build up classification models based on existing SL data to predict novel SL pairs. As a large amount of SL data are available in yeast, various classification models, \emph{e.g.}, decision tree \cite{wong2004combining}, MLE \cite{li2011understanding}, and ensemble classifiers \cite{pandey2010integrative, wu2014silico}, have been studied for yeast SL prediction.

However, existing computational methods for SL prediction have the following limitations when applied to human cancer genome. First, existing supervised learning methods~\cite{wong2004combining,pandey2010integrative,li2011understanding,wu2014silico} were proposed to work on known SL data in yeast instead of human. Without sufficient human SL data for training the models, an alternative way to predict SL for human is to apply the comparative genomics approaches \cite{wu2014silico,deshpande2013comparative,zhan2016towards}. However, human and yeast are evolutionarily distant from each other. Therefore, the comparative analysis between human and yeast for SL prediction may not be reliable. Secondly,  knowledge-based methods ~\cite{kranthi2013identification,jerby2014predicting,jacunski2015connectivity,sinha2017systematic} leverage on the knowledge about the biological networks and other genomic data (\emph{e.g.}, mutation and gene expression profiles). However, they do not employ much information about underlying mechanisms of known SL data in human.

Recently, a comprehensive database called SynLethDB for human SLs becomes publicly available \cite{guo2016synlethdb}. As such, supervised learning methods can be applied on SynLethDB for predicting human SLs. On the other hand, matrix factorization techniques have been successfully applied for various bioinformatics tasks, \emph{e.g.}, PPI prediction \cite{wang2013predicting}, drug-target interaction prediction \cite{liu2016neighborhood}, and drug response prediction~\cite{wang2017improved}.
Motivated by these approaches, in this paper, we have proposed a novel matrix factorization model, named \textsf{SL$^2$MF}, for SL prediction. Differing from existing methods for human SL prediction, \textsf{SL$^2$MF} employs logistic matrix factorization (LMF)~\cite{johnsonlogistic} to model the probability that a pair of genes are likely to have SL interaction. Specifically, a latent vector is assigned to each gene to describe its properties learnt from the data. The probability of two genes to be SL is defined as a logistic function of their latent vectors. To further enhance the prediction accuracy, \textsf{SL$^2$MF} has been extended to integrate the knowledge from PPI networks and Gene Ontology (GO) annotations. This is achieved by exploiting neighborhood regularization to constrain that the learnt latent vectors of genes with similar GO and/or PPI topological properties should be similar in the latent space. We have conducted extensive experiments to evaluate the performance of \textsf{SL$^2$MF}. The experimental results demonstrate the effectiveness of the proposed method and show that using the biological knowledge derived from PPI network and GO can help improve the prediction accuracy. In addition, the comparison between SL$^{2}$MF and DAISY also indicates that SL$^{2}$MF is a useful complement to existing knowledge-based SL prediction methods.

\section{Synthetical Lethality Prediction Model}
\label{section:model}
This section first introduces the notations and problem formulation, and then describes details of the proposed \textsf{SL$^2$MF} model.

\subsection{Preliminary}

In this paper, we denote the set of genes by $ \mathcal{U}= \{u_{i}\}_{i=1}^{m}$, where $m$ is the number of genes. Moreover, we use a binary matrix $\B{Y} \in \mathbb{R}^{m \times m}$, where each element is denoted by $y_{ij} \in \{0, 1\}$, to describe the SL interaction data. If there exists an observed SL interaction between two genes $u_{i}$ and $u_{j}$, we set $y_{ij}$ to 1; otherwise, we set $y_{ij}$ to 0. Note that $\B{Y}$ is symmetric, \emph{i.e.}, $y_{ij}=y_{ji}$. Thus, we treat the gene pairs $(u_{i}, u_{j})$ and $(u_{j}, u_{i})$ as the same pair. Then, we denote the set of all gene pairs by $\mathcal{O}=\{(u_{i}, u_{j})|1\leq i < m, i+1 \leq j\leq m\}$. In other words, we define $\mathcal{O}$ only considering the upper half of $\B{Y}$. In addition, we denote the set of observed SL pairs by $\mathcal{O}^{+}=\{(u_{i}, u_{j})|y_{ij}=1, 1\leq i < m, i+1 \leq j\leq m \}$. The remaining gene pairs in the upper half of $\B{Y}$ are denoted by $\mathcal{O}^{-}=\mathcal{O}\setminus\mathcal{O}^{+}$. We call the pairs in $\mathcal{O}^{-}$ as ``unknown pairs", because there does not exist evidence demonstrating whether these gene pairs are SLs or not.

Moreover, we denote the GO semantic similarities between genes by $\B{S}^{G} \in \mathbb{R}^{m \times m}$, where the $(i, j)$ element $s_{ij}^{G}$ denotes the GO semantic similarity between $u_{i}$ and $u_{j}$. The PPI topological similarities between genes~\footnote{We use genes instead of gene products (\emph{i.e.}, proteins) when we mention their PPI topological similarity.} are denoted by $\B{S}^{P} \in \mathbb{R}^{m \times m}$, where the $(i, j)$ element $s_{ij}^{P}$ is the PPI topological similarity between $u_{i}$ and $u_{j}$. The computation of GO semantic similarities and PPI topological similarities are introduced in Section~\ref{ss:data}. For each gene $u_{i}$, we use $N^{G}(u_{i})$ to denote the set of $k_{1}$ genes that are most similar with $u_{i}$, measured by the GO semantic similarities $\B{S}^{G}$, where $N^{G}(u_{i}) \subset \mathcal{U}\setminus u_{i}$. Similarly, we use $N^{P}(u_{i})$ to denote the set of $k_{2}$ genes that are most similar with $u_{i}$, measured by the PPI topological similarities $\B{S}^{P}$, where $N^{P}(u_{i}) \subset \mathcal{U}\setminus u_{i}$. In this work, $N^{G}(u_{i})$ and $N^{P}(u_{i})$ are called the GO nearest neighbors and PPI nearest neighbors of $u_{i}$, respectively.

\begin{figure}
  \centerline{
    \includegraphics[width = 3.25in]{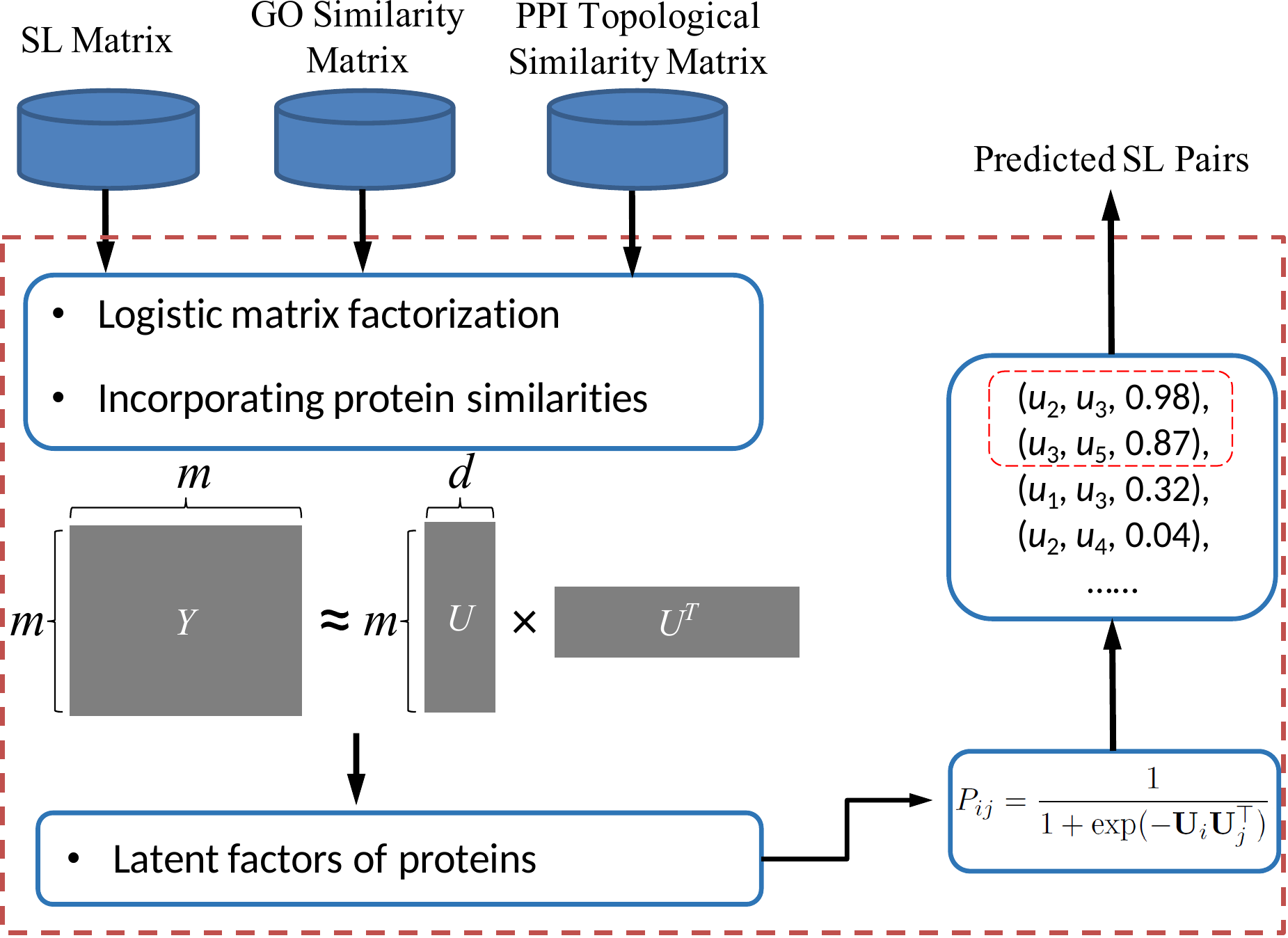}
  }
  \caption{The overall framework of \textsf{SL$^2$MF}. }
  \vspace{-10pt}
  \label{fig:flow}
\end{figure}

The problem studied in this work can be defined as follows: \emph{given a set of observed gene pairs $\mathcal{O}^{+}$ that are known to be SL pairs, how to predict a set of gene pairs that are most likely to form SL interactions from the ``unknown pairs" $\mathcal{O}^{-}$.} We tackle this problem by first predicting the probability that two genes form SL relationship, and then ranking the candidate gene pairs based on the predicted probabilities in descending orders, such that the top-ranked gene pairs are most likely to be SL pairs. Specifically, the method of logistic matrix factorization~\cite{johnsonlogistic,liu2016neighborhood} is utilized to model the SL interaction probabilities between genes. Moreover, neighborhood regularization~\cite{liu2016neighborhood} is used to incorporate both GO semantic similarities and PPI topological similarities between genes to enhance the prediction accuracy. Figure~\ref{fig:flow} shows the overall framework of the proposed \textsf{SL$^2$MF} model.

\subsection{Logistic Matrix Factorization}
The proposed method \textsf{SL$^2$MF} is developed based on logistic matrix factorization, which has been successfully applied in personalized music recommendation~\cite{johnsonlogistic} and drug-target interaction prediction~\cite{liu2016neighborhood}. The objective of logistic matrix factorization is to map genes into a shared low dimensional latent space and model the probabilities of SL interactions between genes using their latent representations. For each gene $u_{i}$, a latent vector $\B{U}_{i} \in \mathbb{R}^{1\times d}$ is used to describe its properties, where $d$ denotes the dimensionality of the latent space, and $d\ll m$. We combine the latent vectors of all genes into a matrix $\B{U}\in \mathbb{R}^{m \times d}$, where $\B{U}_{i}$ is the $i^{th}$ row in $\B{U}$. Then, the SL interaction probability between two genes $u_{i}$ and $u_{j}$ is defined by the following logistic function~\cite{johnsonlogistic}:
\begin{equation}
p_{ij}= \frac{1}{1+\exp(-\B{U}_{i}\B{U}_{j}^{\top})}.
\label{eq:probability}
\end{equation}

To model the SL data, we use the gene pairs in $O^{+}$ that have observed SL interactions as positive training examples, and use the unknown pairs in $\mathcal{O}^{-}$ as negative training examples. Following previous study~\cite{liu2016neighborhood}, we assign higher importance weights to gene pairs that are observed SL interactions. In other words, a known SL pair $(u_{i}, u_{j}) \in \mathcal{O}^{+}$ is treated as $c_{ij}$ $(c_{ij} > 1)$ positive training samples, an unknown gene pair is used as only \textit{one} negative training sample.\footnote{This work is not focusing on developing strategies to weight gene pairs. Thus, we empirically set the weights based on the confidence scores assigned to SLs in SynLethDB (see Section~\ref{ss:importance} for details). Moreover, the proposed \textsf{SL$^2$MF} method is a general framework which can also integrate more sophisticated weighting methods, for example the re-weighted probabilistic models (RPM) proposed in~\cite{wang2017robust}.}
By assuming all the training samples are independent with each other, we can define the likelihood of the observed SL data as follows:
\begin{align}
p(\mathcal{O}|\B{U})
=&\prod_{(u_{i}, u_{j})\in \mathcal{O}^{+}}p_{ij}^{c_{ij}y_{ij}}(1-p_{ij})^{c_{ij}(1-y_{ij})}\nonumber\\
&\prod_{(u_{i}, u_{j}) \in \mathcal{O}^{-}}p_{ij}^{y_{ij}}(1-p_{ij})^{(1-y_{ij})}.
\label{eq:allprobability1}
\end{align}
Note that $c_{ij}(1-y_{ij})=1-y_{ij}$ when $y_{ij}=1$, and $y_{ij}=c_{ij}y_{ij}$ when $y_{ij}=0$. Thus, we can rewrite the likelihood in Eq.\eqref{eq:allprobability1} as follows:
\begin{equation}
  p(\C{O}|\B{U}) = \prod_{i=1}^{m}\prod_{j=i+1}^{m}p_{ij}^{c_{ij}y_{ij}}(1-p_{ij})^{1-y_{ij}}.
  \label{eq:allprobability}
\end{equation}
Zero-mean spherical Gaussian priors are placed on the gene latent vectors as follows~\cite{johnsonlogistic}:
\begin{equation}
p(\B{U}|\sigma^{2}) = \prod_{i=1}^{m}\mathcal{N}(\B{U}_{i}|0, \sigma^{2}\B{I}),
\end{equation}
where $\B{I}$ is the identity matrix, and $\sigma^{2}$ is the parameter used to control the variances of Gaussian distributions. Through Bayesian inference, we have
\begin{equation}
p(\B{U}|\mathcal{O}, \sigma^{2}) \propto p(\mathcal{O}|\B{U})p(\B{U}|\sigma^{2}).
\label{eq:poster}
\end{equation}
The logarithm of the posterior distribution is as follows:
\begin{eqnarray}
&&\log p(\B{U}|\mathcal{O}, \sigma^{2}) \propto \nonumber\\ &&\sum_{i=1}^{m}\sum_{j=i+1}^{m}\bigg[ - (1+c_{ij}y_{ij}-y_{ij})
\log \big(1+\exp(\B{U}_{i}\B{U}_{j}^{\top})\big)\nonumber\\
&&~~~~~~~~~~~~~~+ c_{ij}y_{ij}\B{U}_{i}\B{U}_{j}^{\top} \bigg] -\frac{\lambda}{2}\|\B{U}\|_{F}^{2},
\end{eqnarray}
where $\lambda=\frac{1}{\sigma^{2}}$, and $\|\cdot\|_{F}$ is the Frobenius norm of a matrix. Then, the gene latent vectors can be learned by maximizing the posterior distribution. It is equivalent with minimizing the following loss function:
\begin{align}
L_{\C{O}}=\sum_{i=1}^{m}\sum_{j=i+1}^{m}&\big[(1+c_{ij}y_{ij}-y_{ij})
\log \big(1+\exp(\B{U}_{i}\B{U}_{j}^{\top})\big)\nonumber\\
&- c_{ij}y_{ij}\B{U}_{i}\B{U}_{j}^{\top} \big] +\frac{\lambda}{2}\|\B{U}\|_{F}^{2}.
\label{eq:objective2}
\end{align}

\subsection{Incorporating Gene Similarities}
As mentioned earlier, two types of gene similarities, \emph{i.e.,} GO semantic similarities and PPI topological similarities, are considered to improve the prediction accuracy. Specifically, we assume that genes with similar functional and/or network properties should have similar representations in the latent space. Then, we propose to minimize the following loss function to exploit the GO semantic similarities between genes for SL prediction:
\begin{equation}
L_{\C{G}} = \frac{1}{2}\sum_{i=1}^{m}\sum_{u_{j}\in N^{G}(u_{i})}s^{G}_{ij}\|\B{U}_{i} - \B{U}_{j}\|^{2}_{2}.
\label{eq:objective3}
\end{equation}
Similarly, the PPI similarities between genes can also be incorporated by minimizing the following loss function:
\begin{equation}
L_{\C{P}} = \frac{1}{2}\sum_{i=1}^{m}\sum_{u_{j}\in N^{P}(u_{i})}s^{P}_{ij}\|\B{U}_{i} - \B{U}_{j}\|^{2}_{2}.
\label{eq:objective4}
\end{equation}
Note that the proposed method only considers $k_{1}$ GO nearest neighbors and $k_{2}$ PPI nearest neighbors of each gene $u_{i}$, instead of all its neighbors (i.e., $\mathcal{U} \setminus u_{i}$), to improve the prediction accuracy. Because using all the neighbors may potentially introduce noisy information and thus reduce the model accuracy. In addition, the experimental results in Section~\ref{ss:parameters} also demonstrate that better performance can be achieved by considering only the nearest neighbors of a gene than considering all of its neighbors.

\begin{algorithm2e}
  \caption{The \bf{\textsf{SL$^2$MF}} Algorithm}\label{alg:slmf}
  \LinesNumbered
  \SetKwInOut{KwInput}{Input}
  \SetKwInOut{KwOutput}{Output}
  \BlankLine
  \KwInput{$\B{Y}$, $\B{W}$, $\B{S}^{G}$, $\B{S}^{P}$, $d$, $\lambda$, $\alpha$, $\beta$, $\gamma$, $k_{1}$, $k_{2}$}
  \KwOutput{$\B{U}$}
   Initialize gene latent vectors $\B{U}$ using the Gaussian distribution $\mathcal{N}(0, 1/\sqrt{d})$\;
  Compute the adjacency matrix $\B{A}$ and the Laplacian matrix $\B{L}^{G}$ according to Eq.~\eqref{eq:GOneighbor}\;
  Compute the adjacency matrix $\B{B}$ and the Laplacian matrix $\B{L}^{P}$ according to Eq.~\eqref{eq:TPneighbor}\;
  Set $\varphi_{ik}=0$, $\forall 1\leq i \leq m$ and $1\leq k \leq d$\;
  \For{$iter = 1, 2, \cdots, max\_iter$}{
    Compute the interaction probability matrix $\B{P}$ according to Eq.~\eqref{eq:probability}\;
        \tcp{\small{compute the gradient w.r.t. $\B{U}$}}
        $\B{Z} \leftarrow \big[\B{W} \odot (\B{P}-\B{Y}) + \lambda \B{I} + \alpha \B{L}^{G} + \beta \B{L}^{P}\big] \B{U}$\;
      \For{$i=1, \ldots, m$}{
        \For{$k=1, \ldots, d$}{
        \tcp{\small{$Z_{ik}$ is the $(i,k)$ element in $\B{Z}$}}
            $\varphi_{ik} \leftarrow \varphi_{ik} + Z_{ik}\cdot Z_{ik}$\;
            \tcp{\small{$U_{ik}$ is the $(i,k)$ element in $\B{U}$}}
            $U_{ik} \leftarrow U_{ik} - \gamma \frac{Z_{ik}}{\sqrt{\varphi_{ik}}}$;
        }
      }
 }
\end{algorithm2e}

\subsection{The Unified SL$^2$MF Model}
The final prediction model can be formulated by considering the SL interaction data as well as both GO semantic similarities and PPI topological similarities between genes. By substituting Eq.~\eqref{eq:objective3} and Eq.~\eqref{eq:objective4} into Eq.~\eqref{eq:objective2}, the unified \textsf{SL$^2$MF} model is obtained as follows:
\begin{equation}
  \min_{\B{U}} L_{\C{O}} + \alpha L_{\C{G}} + \beta L_{\C{P}},
  \label{eq:objective5}
\end{equation}
where $\alpha$ and $\beta$ are regularization parameters controlling the influences from GO nearest neighbors and PPI nearest neighbors, respectively.

For simplicity, we define two adjacency matrices $\B{A} \in \mathbb{R}^{m\times m}$ and $\B{B} \in \mathbb{R}^{m\times m}$ to describe the nearest neighbors of genes. The $(i, j)$ element of $\B{A}$ is defined as follows:
\begin{eqnarray}
a_{ij} = \left\{ \begin{array}{rl}
 s_{ij}^{G} & \mbox{if $u_{j}\in N^{G}(u_{i})$}, \\
 0 &\mbox{otherwise.}
       \end{array} \right.
\label{eq:GOneighbor}
\end{eqnarray}
Then, the loss function $L_{\C{G}}$ can be rewritten as follows:
\begin{equation}
  L_{\C{G}} = \frac{1}{2} \mbox{tr}(\B{U}^{\top} \B{L}^{G} \B{U}),
\end{equation}
where $\mbox{tr}(\cdot)$ denotes the trace of a matrix, $\B{L}^{G}=\B{D}^{G} - (\B{A}+\B{A}^{\top})$, $\B{D}_{G} \in \mathbb{R}^{m \times m}$ is a diagonal matrix, in which the diagonal elements are defined as $d^{G}_{ii}=\sum_{j=1}^{m}(a_{ij}+a_{ji})$. Similarly, the $(i, j)$ element of $\B{B}$ is defined as follows:
\begin{eqnarray}
b_{ij} = \left\{ \begin{array}{rl}
 s_{ij}^{P} & \mbox{if $u_{j}\in N^{P}(u_{i})$}, \\
 0 &\mbox{otherwise.}
       \end{array} \right.
\label{eq:TPneighbor}
\end{eqnarray}
We can rewrite the loss function $L_{\C{P}}$ as follows:
\begin{equation}
  L_{\C{P}} = \frac{1}{2}\mbox{tr}(\B{U}^{\top} \B{L}^{P} \B{U}),
\end{equation}
where $\B{L}^{P}=\B{D}^{P} - (\B{B}+\B{B}^{\top})$. $\B{D}^{P} \in \mathbb{R}^{m \times m}$ is a diagonal matrix, in which the diagonal elements are defined as $d^{P}_{ii}=\sum_{j=1}^{m}(b_{ij}+b_{ji})$. Then, the unified model in Eq.~\eqref{eq:objective5} can be rewritten as follows:
\begin{align}
 \min_{\B{U}}&\frac{1}{2}\sum_{i=1}^{m}\sum_{j=1}^{m} w_{ij} \big[\log \big(1+\exp(\B{U}_{i}\B{U}_{j}^{\top})\big) - y_{ij}\B{U}_{i}\B{U}_{j}^{\top}\big] \nonumber\\ &+\frac{\lambda}{2}\|\B{U}\|_{F}^{2}
 +\frac{\alpha}{2} \mbox{tr}(\B{U}^{\top} \B{L}^{G} \B{U})
 +\frac{\beta}{2}\mbox{tr}(\B{U}^{\top} \B{L}^{P} \B{U}),
 \label{eq:objective6}
\end{align}
where $w_{ij}$ is defined as follows:
\begin{eqnarray}
w_{ij} = \left\{ \begin{array}{rl}
 c_{ij} & \mbox{if $y_{ij}=1$ and $i\neq j$}, \\
 1 & \mbox{if $y_{ij}=0$ and $i\neq j$}, \\
 0 &\mbox{if $i=j$}.
       \end{array} \right.
\label{eq:weight}
\end{eqnarray}
Note that $y_{ij}=y_{ji}$ and $c_{ij}=c_{ji}$, thus $w_{ij}=w_{ji}$.

The problem in Eq.~\eqref{eq:objective6}, denoted by $L$, can be solved by the gradient descent optimization procedure. The gradient of Eq.~\eqref{eq:objective6} with respect to $\B{U}$ is as follows:
\begin{equation}\label{eq:gradient}
  \frac{\partial L}{\partial \B{U}} = \big[\B{W} \odot (\B{P}-\B{Y}) + \lambda \B{I} + \alpha \B{L}^{G} + \beta \B{L}^{P}\big] \B{U},
\end{equation}
where $\odot$ is the Hadamard product of two matrices. $\B{W}\in \mathbb{R}^{m \times m}$ is the weighting matrix, where the $(i, j)$ element is $w_{ij}$ (refer to Eq.~\eqref{eq:weight}). $\B{P} \in \mathbb{R}^{m \times m}$ is the interaction probability matrix, in which the $(i, j)$ element is $p_{ij}$ (refer to Eq.~\eqref{eq:probability}). Following~\cite{johnsonlogistic}, we adopt the AdaGrad algorithm~\cite{duchi2011adaptive} to accelerate the convergence of the gradient descent procedure. The details of the optimization algorithm developed for \textsf{SL$^2$MF} are summarized in Algorithm~\ref{alg:slmf}.

Once having learned the model parameters $\B{U}$, Eq~\eqref{eq:probability} is used to compute the probability that a candidate pair of genes are SL. Then, all candidate gene pairs $\mathcal{O}^{-}$ are ranked based on the predicted probabilities in descending orders, and the top-ranked gene pairs will be chosen as the predictions.


\subsection{Discussions}
As shown in Eq.~\eqref{eq:objective5} and Eq.~\eqref{eq:objective6}, the unified model is a linear combination of the optimization problem in Eq.~\eqref{eq:objective2} and the neighborhood regularization constraints Eq.~\eqref{eq:objective3} and Eq.~\eqref{eq:objective4}. Therefore, the construction of the unified model does not fully follow the spirit of probabilistic models. However, inspired by previous research work about social recommendation~\cite{liu2017learning}, we can note that the usage of the regularization terms $\frac{\lambda}{2}\|\B{U}\|_{F}^{2}
 +\frac{\alpha}{2} \mbox{tr}(\B{U}^{\top} \B{L}^{G} \B{U})
 +\frac{\beta}{2}\mbox{tr}(\B{U}^{\top} \B{L}^{P} \B{U})$ in Eq.~\eqref{eq:objective6} is equivalent to assigning the following priors on the gene latent vectors $\B{U}$:
\begin{align}
  p(\B{U}|\B{L}^{G}, &\B{L}^{P}, \lambda, \alpha, \beta) \propto \nonumber \\
  &\C{MN}_{m \times d} \bigg(\B{0}, (\lambda \B{I} + \alpha \B{L}^{G} + \beta \B{L}^{P})^{-1}, \B{I} \bigg),
  \label{eq:mvnprior}
\end{align}
where $\C{MN}_{a \times b}(\B{M}, \B{\Sigma}, \B{\Theta})$ is a matrix variate normal (MVN) distribution\footnote{The density function for a random matrix $\B{X}$ following the MVN distribution $\mathcal{MN}_{a \times b}(\B{M}, \B{\Sigma}, \B{\Theta})$ is as follows:
\begin{equation}
 p(\B{X})=\frac{\exp(-\frac{1}{2}\mbox{tr}\big[\B{\Theta}^{-1}(\B{X}-\B{M})^{\top}\B{\Sigma}^{-1}(\B{X}-\B{M})\big])}{(2\pi)^{ab/2}|\B{\Theta}|^{a/2}|\B{\Sigma}|^{b/2}}.\nonumber
\end{equation}
}~\cite{gupta1999matrix} with the mean $\B{M} \in \mathbb{R}^{a \times b}$, row covariance $\B{\Sigma} \in \mathbb{R}^{a \times a}$, and column covariance $\B{\Theta} \in \mathbb{R}^{b \times b}$. In Eq.~\eqref{eq:mvnprior}, the relations between different rows of $\B{U}$ are described by the row precision matrix: $\B{\Omega}=\lambda \B{I} + \alpha \B{L}^{G} + \beta \B{L}^{P}$. Note that each row of $\B{U}$ denotes the latent vector of an individual gene. Therefore, in other words, the relations between different genes are modeled by the row precision matrix $\B{\Omega}$. Through Bayesian inference, we have
\begin{equation}
  p(\B{U}|\mathcal{O}, \B{L}^{G}, \B{L}^{P}, \lambda, \alpha, \beta) \propto p(\mathcal{O}|\B{U})p(\B{U}|\B{L}^{G}, \B{L}^{P}, \lambda, \alpha, \beta).
\end{equation}
The gene latent vectors $\B{U}$ can be learned by maximizing the above posterior distribution, which is equivalent to solving the optimization problem in Eq.~\eqref{eq:objective6}.

\section{Results}
\label{section:results}

In this section, we first introduce the data used in the experiments and the experimental settings. Then, we show the performances of SL$^{2}$MF under different settings.

\subsection{Data and Experimental Settings}
\label{ss:data}
The proposed \textsf{SL$^2$MF} method works on the SL interaction matrix and various gene similarity matrices, including PPI topological similarity matrix and GO semantic similarity matrix. The SL dataset was downloaded from the database of SynLethDB \cite{guo2016synlethdb}. SynLethDB integrates SL interaction data from four different sources: (1) SL pairs manually curated \cite{li2014syn}, (2) SL pairs extracted by text mining \cite{guo2016synlethdb}, (3) genetic interactions detected by GenomeRNAi \cite{schmidt2013genomernai} and shRNA (i.e., the DECIPHER project\footnote{http://www.decipherproject.net/shRNA-libraries/bi-specific/}), and (4) SL pairs computationally predicted by DAISY \cite{jerby2014predicting}. Overall, there are 19,667 SL pairs involving 6,375 genes in the SL matrix. The sparsity of the SL interaction matrix is $99.90\%$, thus the SL prediction is a very challenging prediction task. In SynLethDB, a confidence score is assigned to each SL pair, considering the accuracies of different identification methods. These confidence scores have been exploited to define the weights assigned to SL pairs in \textsf{SL$^2$MF} (refer to Table~\ref{t:weights} for details). Moreover, we downloaded the HPRD database \cite{keshava2009human} to construct the PPI topological similarity matrix. FSWeight \cite{chua2006exploiting}, which is a topological similarity between genes based on the number of common neighbors in the PPI data, was then calculated. Among various methods for computing the GO term similarity, we used the method proposed in~\cite{wang2007new} and then calculated the GO semantic similarity between genes. It is known that GO has three sub-ontologies, namely biological process (BP), molecular function (MF) and cellular component (CC). In GO (version: June 2017), there are 29,660 terms in BP, 11,120 terms in MF, and 4,115 terms in CC, respectively. BP has many more terms and are more enriched than MF and CC. Therefore, we computed the semantic similarity between genes only using their BP terms.


In our experiments, we adopted 5-fold cross-validation to evaluate the proposed method. The known SL pairs $\C{P}$ in SynLethDB were equally divided into 5 non-overlapping subsets (\emph{i.e.}, $\C{P}_{1}$, $\C{P}_{2}$, $\C{P}_{3}$, $\C{P}_{4}$, $\C{P}_{5}$). In each round, a subset $\C{P}_{i} (1\leq i\leq5)$ of SL interactions were chosen for model testing, and the other four subsets were used as positive examples for model training (i\emph{.e.}, $\C{O}^{+}$). The objective of \textsf{SL$^2$MF} is to predict the potential SL pairs in $\C{O}^{-}$ (\emph{i.e.}, $\C{P}_{i}$) via ranking the gene pairs in $\C{P}_{i}$ before other gene pairs in $\C{O}^{-}$.
Therefore, we use AUC (\emph{i.e.}, area under the ROC curve) and AUPR (\emph{i.e.} area under the precision-recall curve) as evaluation metrics. In particular, the AUC and AUPR scores were calculated based on the predictions of all candidate gene-gene pairs.

The proposed SL prediction model is built based on the known SL interaction data in human, the GO semantic similarities and PPI topological similarities between genes. Therefore, existing supervised learning methods~\cite{wong2004combining,pandey2010integrative,li2011understanding,wu2014silico} for yeast SL prediction
and other knowledge-based methods~\cite{kranthi2013identification,jerby2014predicting,jacunski2015connectivity,sinha2017systematic} for human SL prediction cannot be directly used as baselines in this study. As SL prediction shares similar spirits with drug-target interaction prediction~\cite{liu2016neighborhood}, we adopt the similarity-based drug-target interaction prediction method BLM-NII~\cite{mei2012drug} as the baseline method. In BLM-NII, the linear combination weight $\alpha$ was chosen from $\{0, 0.1, 0.2, \cdots, 1.0\}$, and the \emph{max} function was used to integrate the interaction scores predicted independently by using the GO semantic similarities and the PPI topological similarities. For \textsf{SL$^2$MF}, we empirically set the dimensionality of latent space $d$ to 50. We assigned uniform weights to SL pairs as shown in Table~\ref{t:weights}, and the parameter $c$ was set to 50. The regularization parameters $\lambda$, $\alpha$, and $\beta$ were set to $0.01$, $1.0$, and $10$, respectively. The learning rate $\gamma$ of the optimization algorithm was set to $2^{-5}$. The number of GO semantic nearest neighbors and PPI topological nearest neighbors $k_{1}$ and $k_{2}$ were set to $100$. In Section~\ref{ss:parameters}, we will discuss the impacts of these parameters.

\label{ss:cvresults}
\begin{table}
\caption{The 5-fold cross-validation AUC and AUPR scores of BLM-NII and \textsf{SL$^2$MF} with respect to different settings.} \label{t:cv}
\centering
\small
\begin{tabular}{|c|c|c|}\hline
Method    & AUC  & AUPR\\\hline
BLM-NII   & 0.7228$\pm$0.0293 & 0.0011$\pm$0.0010 \\ \hline
SL Matrix & 0.8051$\pm$0.0060 & 0.2081$\pm$0.0072 \\ \hline
SL+PPI    & 0.8330$\pm$0.0029 & 0.1326$\pm$0.0046 \\ \hline
SL+GO     & 0.8437$\pm$0.0049 & 0.2414$\pm$0.0062 \\ \hline
SL+PPI+GO & 0.8480$\pm$0.0048 & 0.2388$\pm$0.0057 \\ \hline
\end{tabular}
\vspace{-10pt}
\end{table}

\subsection{5-fold Cross Validation Results}

Table~\ref{t:cv} shows the AUC and AUPR scores of BLM-NII and \textsf{SL$^2$MF} with respect to different inputs. In Table~\ref{t:cv}, ``SL Matrix" refers to \textsf{SL$^2$MF} without using any similarity matrices, ``SL+PPI" refers to \textsf{SL$^2$MF} incorporating only the PPI topological similarity matrix via setting the parameter $\alpha$ to 0 in Eq.~\eqref{eq:objective6}, and ``SL+GO" refers to \textsf{SL$^2$MF} incorporating only the GO semantic similarity matrix by setting $\beta$ to 0. Similarly, ``SL+PPI+GO" indicates that \textsf{SL$^2$MF} works on SL matrix together with both the PPI topological similarity matrix and the GO semantic similarity matrix. Based on the results in Table~\ref{t:cv}, we can draw the following two conclusions.

First, both PPI topological similarities and GO similarities can help improve the performance for predicting SL interactions, in terms of AUC. For example, the AUC scores are improved by 3.47\% and 4.79\% after incorporating the PPI topological similarities and GO similarities, respectively. Moreover, the GO similarities can also help improve AUPR scores. However, by incorporating the PPI similarities into the model, AUPR scores are dropped off. This is to be expected, since the strategy used to improve AUC scores is not guaranteed to improve AUPR scores, and also because AUPR punishes highly ranked false positives much more than AUC~\cite{davis2006relationship}. One potential explanation for this observation is as follows. When incorporating PPIs, less non-SL gene pairs are ranked before SL gene pairs, thus increasing AUC. However, these falsely ranked non-SL gene pairs may be ranked at higher positions, thus decreasing AUPR.

Secondly, the GO semantic similarity matrix performs better than the PPI topological similarity matrix in improving the prediction accuracy. Compared with the PPI topological similarity matrix, the GO semantic similarity matrix is able to further improve the AUC and AUPR by 1.28\% and 82.05\%, respectively. One reason could be that the PPI topological similarity matrix is much sparser than the GO semantic similarity matrix. In particular, the sparsity of the PPI topological similarity matrix $\B{S}^{P}$ and GO semantic similarity matrix $\B{S}^{G}$ are 98.65\% and 17.04\%, respectively. The sparsity is defined as the percentage of zero elements in the matrix.

Thirdly, as shown in Table~\ref{t:cv}, we can notice that the proposed \textsf{SL$^2$MF} method significantly outperforms BLM-NII, in terms of AUC and AUPR, and the AUPR score achieved by BLM-NII is around 0.001. The potential reason is that both similarity matrices $\B{S}^{G}$ and $\B{S}^{P}$ are sparse. Thus, BLM-NII cannot exploit enough gene similarities for accurate SL prediction. In addition, this result also demonstrates the effectiveness of \textsf{SL$^2$MF} in handling sparse information for SL prediction.


\begin{table}
\caption{Different definitions of the importance weights assigned to SL pairs. Here $\varepsilon_{ij}$ is the confidence score of each SL pair in SynLethDB~\cite{guo2016synlethdb}.} \label{t:weights}
\small
\centering
\begin{tabular}{|l|l|}\hline
    & Definitions \\\hline
Uniform Weights    & $c_{ij}=c$  \\ \hline
Linear Weights     & $c_{ij} = 1+c \varepsilon_{ij}$ \\ \hline
Loglinear Weights  & $c_{ij} = 1+c \log(1+\varepsilon_{ij})$ \\\hline
\end{tabular}
\vspace{-12pt}
\end{table}

\subsection{Benefits of Importance Weights}
\label{ss:importance}

The SL pairs collected in SynLethDB are usually supported by different types of evidences. They are more trustworthy and important than the unknown pairs. Hence, we design a parameter $c_{ij}$ in Eq.~\eqref{eq:allprobability} to control the importance levels for known SL pairs. In particular, each SL pair is treated as $c_{ij}$ positive training instances while each unknown pair is treated as \emph{\textbf{a single}} negative instance. In this paper, we have studied three different definitions of the importance weights $c_{ij}$ assigned to SL pairs based on the confidence scores defined in SynLethDB~\cite{guo2016synlethdb}, as shown in Table~\ref{t:weights}.

\begin{figure}
  \centerline{
    \includegraphics[width = 3.25in]{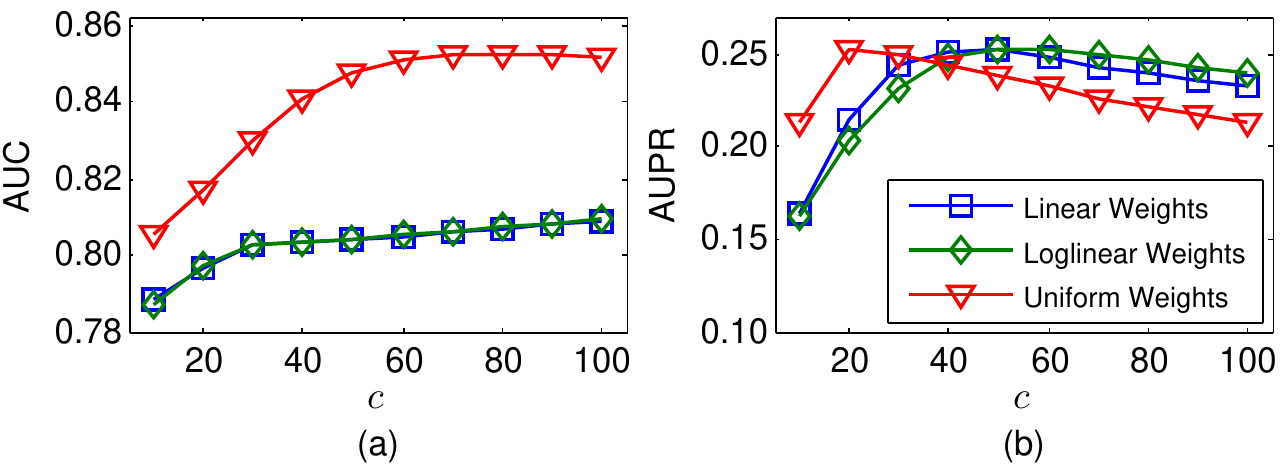}
  }
  \vspace{-8pt}
  \caption{The AUC and AUPR scores of \textsf{SL$^2$MF} with different definitions of the weight $c_{ij}$ for ``SL+GO+PPI". }\label{fig:c}
  \vspace{-10pt}
\end{figure}

Figure~\ref{fig:c} shows the performances of the proposed \textsf{SL$^2$MF} method with respect to different settings of $c_{ij}$. As shown in Figure~\ref{fig:c}, better performances can usually be achieved by assigning uniform weights to SL pairs. This observation is to be expected. Because the calculation of AUC and AUPR scores does not consider the quality of the SL pairs. Uniform weights may achieve better performance than linear weights and log-linear weights, in terms of AUC and AUPR. Moreover, the confidence scores provided in SynLethDB are empirically set by the authors of SynLethDB. These scores may not be optimal for SL prediction tasks. For uniform weights, when $c$ is set to 1 (\emph{i.e.}, known SL pairs are equally important as unknown pairs), the performance of \textsf{SL$^2$MF} is poor, \emph{i.e.}, the AUC is only 0.5119. When we increase the value of $c$, the performance of \textsf{SL$^2$MF} is significantly improved. However, when $c$ is large enough (\emph{e.g.}, $c > 50$), the performance of \textsf{SL$^2$MF} becomes stable. As such, we recommend that 50 is a proper setting for $c$ when predicting SL interactions by incorporating both similarities.

\subsection{Parameter Sensitivity Analysis}
\label{ss:parameters}

This sections focuses on sensitivity analysis for parameters $d$, $\alpha$, $\beta$, $\lambda$, $k_1$, and $k_{2}$. Note that we use SL matrix and both similarity matrices (\emph{i.e.}, ``SL+GO+PPI") to show the effects of parameters $d$, $\alpha$, and $\beta$. For parameters $k_1$ and $k_{2}$, we show their settings for ``SL+PPI" and ``SL+GO", respectively.

\begin{figure}
  \centerline{
    \includegraphics[width = 3.25in]{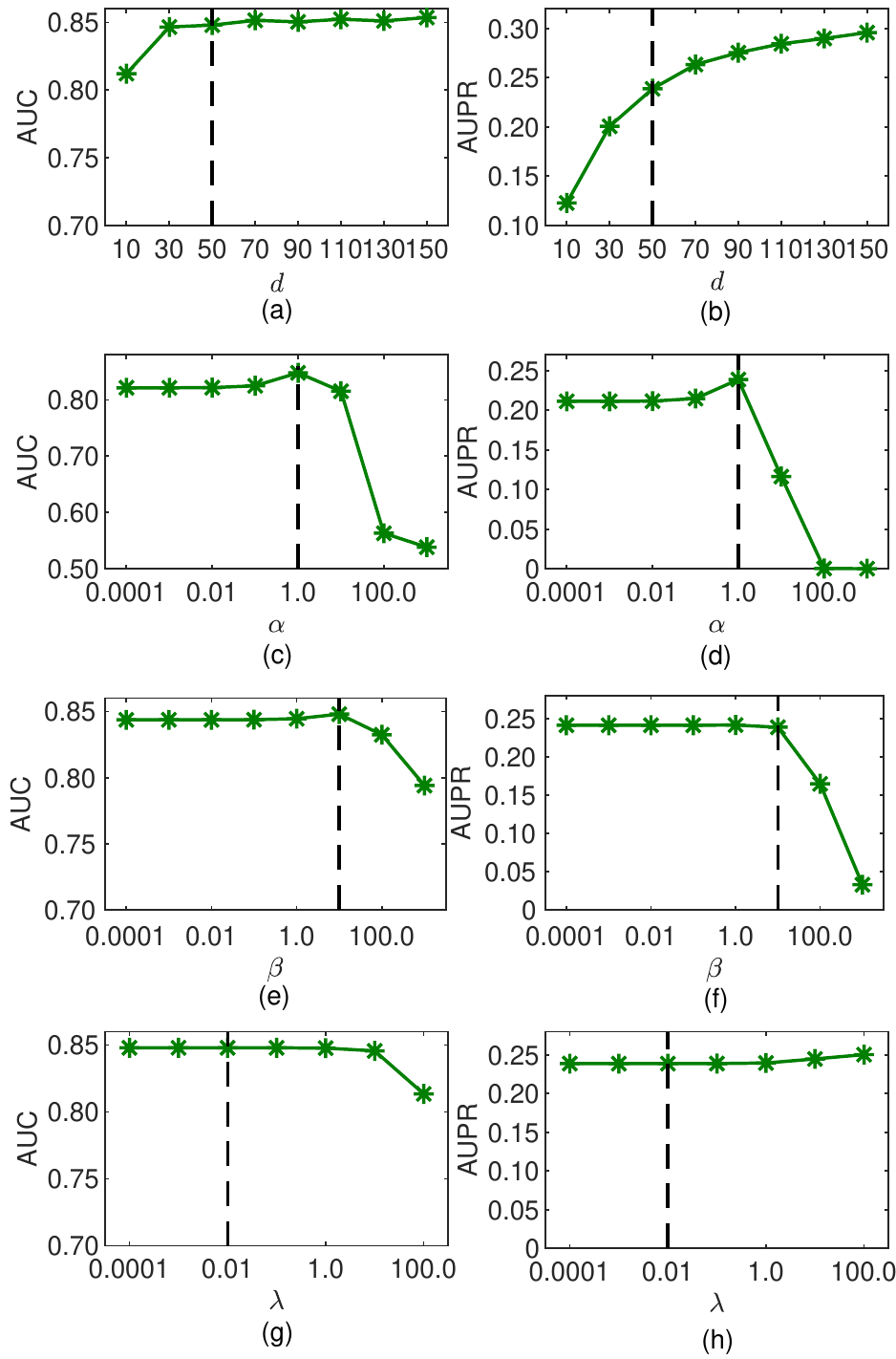}
  }
  \vspace{-5pt}
  \caption{Impacts of $d$, $\alpha$, $\beta$, and $\lambda$ for ``SL+GO+PPI". }\label{fig:Paras}
  \vspace{-10pt}
\end{figure}

Parameter $d$ is the dimensionality of the learned latent vectors of genes. As shown in Figure~\ref{fig:Paras} (a), AUC becomes stable when $d \geq 30$. Moreover, AUPR generally improves with the increase of $d$. However, larger $d$ causes more computation time used to learn gene latent vectors. Considering both the efficiency and accuracy, we empirically set $d$ to 50 in this study. Parameters $\alpha$  and $\beta$ are coefficients controlling contributions of GO semantic similarity matrix and PPI topological similarity matrix, respectively. Figures~\ref{fig:Paras} (c) and \ref{fig:Paras} (d) show AUC and AUPR scores of proposed method with respect to different settings of $\alpha$, by fixing $\beta$ to 10. We can observe that the optimal value for $\alpha$ is 1. In Figures \ref{fig:Paras} (e) and (f), we fix $\alpha$ to 1 and then vary the values of $\beta$. We can observe that both AUC and AUPR are consistently high when $\beta \in [0.0001, 10]$, but they are decreased when $\beta$ is further increased to 100 and 1000. Gradually decreasing $\beta$ from 10 to 0.0001 will not affect the performance, indicating that the PPI topological similarity matrix has less impact on the performance than the GO semantic similarity matrix. This result is also consistent with Table~\ref{t:cv}, where ``SL+GO" performs better than ``SL+PPI". Moreover, Figures (g) and (f) summarize the performances with respect to different settings of $\lambda$. We can notice that both AUC and AUPR are very stable when $\lambda \in [0.0001, 10]$. However, when $\lambda$ is further increased, AUC is dramatically decreased and AUPR is slightly increased.

\begin{figure}
  \centerline{
    \includegraphics[width = 3.25in]{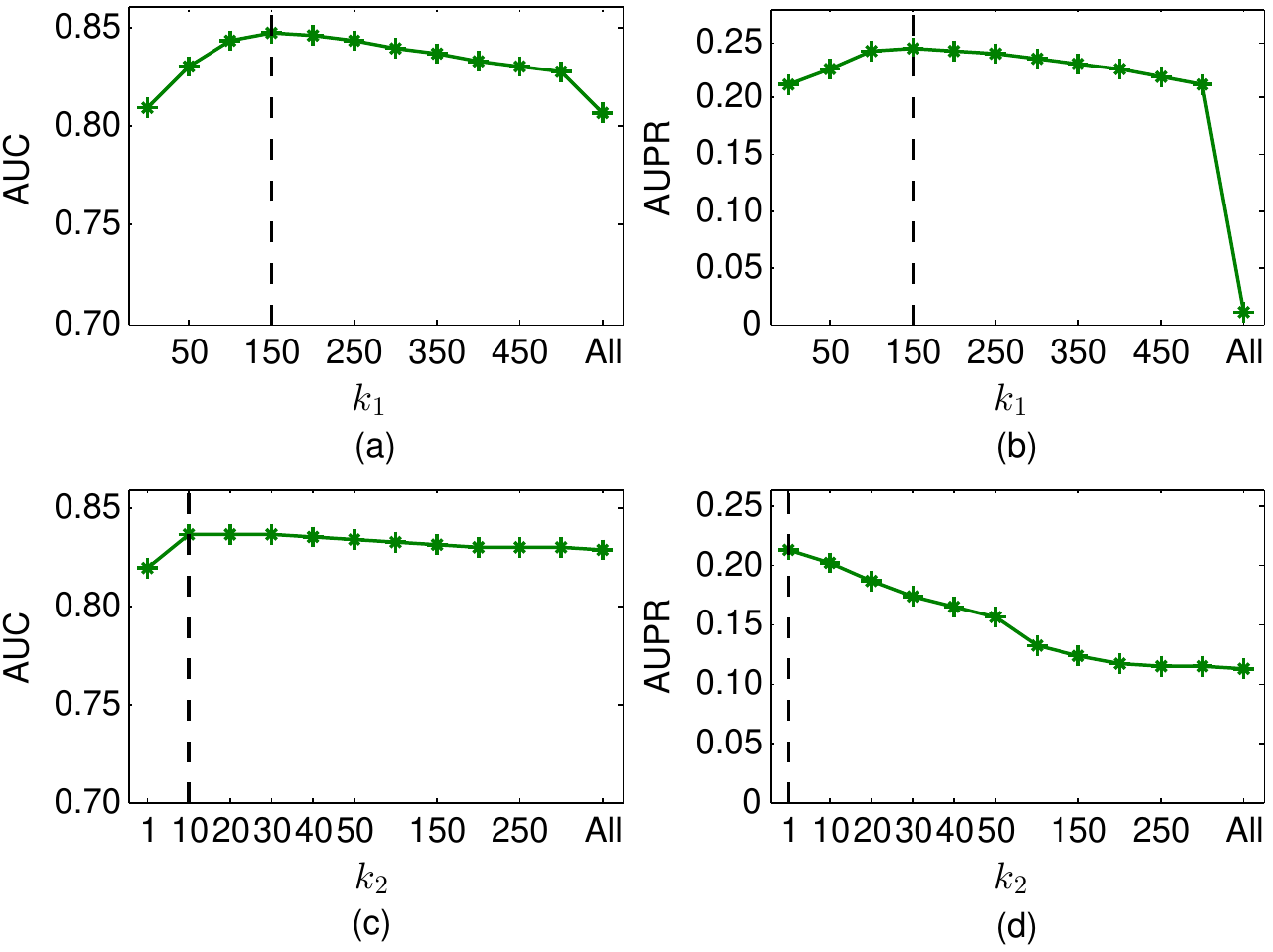}
  }
  \vspace{-5pt}
  \caption{Impacts of $k_{1}$ for ``SL+GO" and $k_{2}$ for ``SL+PPI". }\label{fig:neigbors}
  \vspace{-10pt}
\end{figure}

Parameters $k_1$ and $k_{2}$ are the numbers of nearest neighbors used in regularization constraints Eq.~\eqref{eq:objective3} and Eq.~\eqref{eq:objective4}. For ``SL+GO", it is clear that the optimal value of $k_1$ is 150 as shown in Figures \ref{fig:neigbors} (a) and (b). For ``SL+PPI", the number of nearest neighbors utilized by regularization is supposed to be small, due to the high sparsity of the PPI topological similarity matrix. Figure \ref{fig:neigbors} (c), where the optimal value of $k_2$ is 10, also confirms this supposition. Overall, the recommended settings of parameters $k_1$ and $k_{2}$ are 150 and 10 for ``SL+GO" and ``SL+PPI", respectively. 
Moreover, the rightmost points in Figure~\ref{fig:neigbors} refer to AUC and AUPR scores using ``\emph{All}" the neighbors (\emph{i.e.}, 6,374 neighbors). Therefore, better performances can be achieved by exploiting only a few nearest neighbors instead of all the neighbors.

\subsection{Comparison with DAISY}

As mentioned earlier, \textsf{SL$^2$MF} is a supervised matrix factorization model, whereas DAISY~\cite{jerby2014predicting} is a well-known knowledge-based method for SL prediction which is based on three hypotheses about SL gene pairs. More importantly, we have conducted experiments and obtained all the results above using the whole SynLethDB dataset comprising 19,667 SL pairs, including 5,740 SL pairs predicted by DAISY. To fairly compare the predictions of \textsf{SL$^2$MF} with those of DAISY, we modify the data setting for \textsf{SL$^2$MF} as follows.

\begin{figure}
  \centerline{
    \includegraphics[width = 3.25in]{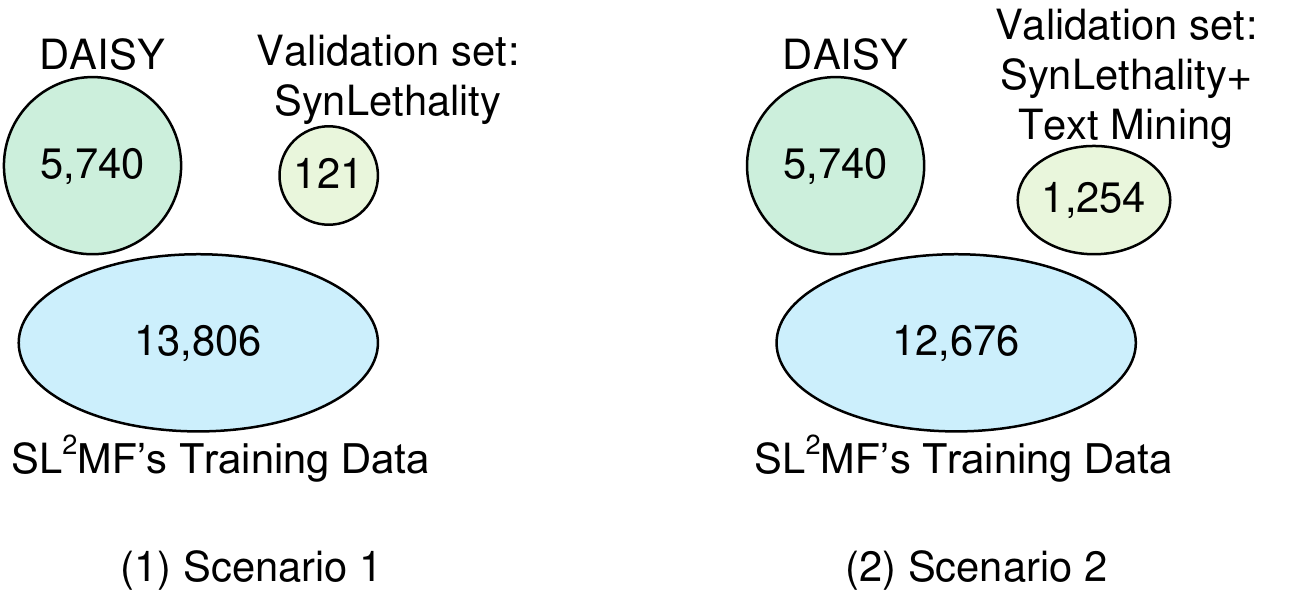}
  }
  \caption{Two different scenarios to run \textsf{SL$^2$MF} for comparison with DAISY. The set of SLs predicted by DAISY and the validation SL set have no overlap in scenario 1, and these two sets have 3 common SL pairs in scenario 2. Both scenarios have 19,667 SL pairs.}\label{fig:daisy}
  \vspace{-10pt}
\end{figure}

We first remove the SL pairs predicted by DAISY from SynLethDB. Then, we also reserve a validation set consisting of SL pairs with high reliability (e.g., high confidence scores in SynLethDB). After excluding these two parts, the remaining SL pairs in SynLethDB will be used as positive training samples to train \textsf{SL$^2$MF}. In particular, we have worked on two scenarios for comparing \textsf{SL$^2$MF} and DAISY as shown in Figure \ref{fig:daisy}. In scenario 1, the manually curated SL pairs (\emph{i.e.}, the SynLethality database~\cite{li2014syn}  which is also the part of SynLethDB with the highest confidence scores) are used as validation set. In scenario 2, the validation set is expanded by further including those SL pairs extracted by text mining~\cite{guo2016synlethdb}.

Once completing the model training of \textsf{SL$^2$MF}, the candidate SLs are ranked based on the predicted probabilities. The validation AUC scores of \textsf{SL$^2$MF} are 0.7330 in scenario 1 and 0.6701 in scenario 2. The AUC score in scenario 2 is relatively lower, probably because less SL pairs are used to train \textsf{SL$^2$MF} in the second scenario, which might affect the performance of the trained \textsf{SL$^2$MF} model. However, in both scenarios, the validation AUPR scores of \textsf{SL$^2$MF} are less than 0.005. One potential reason is that AUPR punishes highly ranked false positives much more than AUC~\cite{davis2006relationship}, and the validation SLs do not rank high based on the interaction probabilities predicted by \textsf{SL$^2$MF}.


Moreover, we have also studied the top 5,740 SL pairs (i.e., the same size of DAISY) predicted by \textsf{SL$^2$MF}. 
In the first scenario, 14 out of 5,740 pairs predicted by \textsf{SL$^2$MF} are also in the SynLethality database, while DAISY has no overlap with SynLethality database. In the second scenario, 27 out of 5,740 pairs predicted \textsf{SL$^2$MF} are verified by the validation set (\emph{i.e.}, SynLethality + Text Mining), while only 3 SL pairs in DAISY are verified by text mining. In addition, we also observed that \textsf{SL$^2$MF} and DAISY have limited overlap in their predictions. In scenario 1, they have only 5 predicted SL pairs in common: (POLR2A, WRAP53), (PARP1, PRKDC), (EGFR, IGFBP3), (POLD1, POLR2A), (CHEK1, MRE11); in scenario 2, only 3 SL pairs are commonly predicted by both methods: (POLR2A, WRAP53), (EGFR, IGFBP3), (POLD1, POLR2A). Such a small overlap actually makes sense as \textsf{SL$^2$MF} and DAISY have different prediction mechanisms and use different data sources for prediction. 
For example, in scenario 1, the 5,740 SL pairs predicted by \textsf{SL$^2$MF} correspond to 3,555 genes. Meanwhile, the 5,740 SL pairs predicted by DAISY involve 3,796 genes. However, there are only 998 genes covered by both DAISY and \textsf{SL$^2$MF} predictions. Given \textsf{SL$^2$MF}'s good performance evaluated by cross validation, we believe that \textsf{SL$^2$MF} is a useful complement to the existing methods, such as DAISY. 
%
%

\section{Conclusion and Future Work}
\label{section:conclusion}
In this paper, we have proposed a novel method for predicting the genetic interactions of synthetic lethality, named \textsf{SL$^2$MF}, which exploits logistic matrix factorization to model the probability that two genes are likely to form synthetic lethality. Because the observed SL pairs are supported by different types of evidence, we proposed to assign higher importance weights to known SL pairs, and assigned lower importance weights to other gene pairs. To further enhance the prediction accuracy of \textsf{SL$^2$MF}, we integrated both GO semantic similarities and PPI topological similarities between genes into learning gene latent vectors. Extensive experiments have been performed to evaluate the prediction accuracies of \textsf{SL$^2$MF}, under different settings. Experimental results have shown that \textsf{SL$^2$MF} is able to achieve promising performance for SL prediction. We believe that it can serve as a good baseline for future studies on this topic.

The future work will focus on the following directions. First, we plan to integrate more sophisticated weighting method, \emph{e.g.}, the re-weighted probabilistic model (RPM) proposed in~\cite{wang2017robust}, to improve the prediction accuracy of \textsf{SL$^2$MF}. Moreover, the lack of a nice independent validation dataset, which covers all required information of various methods, to compare \textsf{SL$^2$MF} with existing SL prediction methods is a weakness of this work.
To address this issue, it will be a highly desirable future work to expand the current version of SynLethDB database to incorporate SLs identified by siRNA/CRISPR based knockdown screens or other existing methods.
Furthermore, we also would like to integrate more types of data (\emph{e.g.}, protein domains, TCGA data) to further improve \textsf{SL$^2$MF}'s performance. As more data are processed, we can also investigate feature-based classification models (\emph{e.g.}, Gradient Boosting Machines \cite{chen2016xgboost}) for SL prediction. Last but not least, verifying the effectiveness of \textsf{SL$^2$MF} through wet-lab experiments is also a potential research direction for future work.

\section{Acknowledgement}
This research is supported, in part, by the National Research Foundation, Prime Minister's Office, Singapore under its IDM Futures Funding Initiative, and the Alibaba-NTU Singapore Joint Research Institute. This research is also supported, in part, by the Start-up grant of ShanghaiTech University.


\bibliographystyle{IEEEtran}
\bibliography{references}    

\begin{thebibliography}{10}
\providecommand{\url}[1]{#1}
\csname url@samestyle\endcsname
\providecommand{\newblock}{\relax}
\providecommand{\bibinfo}[2]{#2}
\providecommand{\BIBentrySTDinterwordspacing}{\spaceskip=0pt\relax}
\providecommand{\BIBentryALTinterwordstretchfactor}{4}
\providecommand{\BIBentryALTinterwordspacing}{\spaceskip=\fontdimen2\font plus
\BIBentryALTinterwordstretchfactor\fontdimen3\font minus
  \fontdimen4\font\relax}
\providecommand{\BIBforeignlanguage}[2]{{%
\expandafter\ifx\csname l@#1\endcsname\relax
\typeout{** WARNING: IEEEtran.bst: No hyphenation pattern has been}%
\typeout{** loaded for the language `#1'. Using the pattern for}%
\typeout{** the default language instead.}%
\else
\language=\csname l@#1\endcsname
\fi
#2}}
\providecommand{\BIBdecl}{\relax}
\BIBdecl

\bibitem{ashworth2011genetic}
A.~Ashworth, C.~J. Lord, and J.~S. Reis-Filho, ``Genetic interactions in cancer
  progression and treatment,'' \emph{Cell}, vol. 145, no.~1, pp. 30--38, 2011.

\bibitem{hartwell1997integrating}
L.~H. Hartwell, P.~Szankasi, C.~J. Roberts, A.~W. Murray, and S.~H. Friend,
  ``Integrating genetic approaches into the discovery of anticancer drugs,''
  \emph{Science}, vol. 278, no. 5340, pp. 1064--1068, 1997.

\bibitem{mclornan2014applying}
D.~P. McLornan, A.~List, and G.~J. Mufti, ``Applying synthetic lethality for
  the selective targeting of cancer,'' \emph{New England Journal of Medicine},
  vol. 371, no.~18, pp. 1725--1735, 2014.

\bibitem{simons2001establishment}
A.~Simons, N.~Dafni, I.~Dotan, Y.~Oron, and D.~Canaani, ``Establishment of a
  chemical synthetic lethality screen in cultured human cells,'' \emph{Genome
  research}, vol.~11, no.~2, pp. 266--273, 2001.

\bibitem{turner2008synthetic}
N.~C. Turner, C.~J. Lord, E.~Iorns, R.~Brough, S.~Swift, R.~Elliott, S.~Rayter,
  A.~N. Tutt, and A.~Ashworth, ``A synthetic lethal sirna screen identifying
  genes mediating sensitivity to a parp inhibitor,'' \emph{The EMBO journal},
  vol.~27, no.~9, pp. 1368--1377, 2008.

\bibitem{luo2009genome}
J.~Luo, M.~J. Emanuele, D.~Li, C.~J. Creighton, M.~R. Schlabach, T.~F.
  Westbrook, K.-K. Wong, and S.~J. Elledge, ``A genome-wide rnai screen
  identifies multiple synthetic lethal interactions with the ras oncogene,''
  \emph{Cell}, vol. 137, no.~5, pp. 835--848, 2009.

\bibitem{martins2015linking}
M.~M. Martins, A.~Y. Zhou, A.~Corella, D.~Horiuchi, C.~Yau, T.~Rakshandehroo,
  J.~D. Gordan, R.~S. Levin, J.~Johnson, J.~Jascur \emph{et~al.}, ``Linking
  tumor mutations to drug responses via a quantitative chemical--genetic
  interaction map,'' \emph{Cancer discovery}, vol.~5, no.~2, pp. 154--167,
  2015.

\bibitem{du2017genetic}
D.~Du, A.~Roguev, D.~E. Gordon, M.~Chen, S.-H. Chen, M.~Shales, J.~P. Shen,
  T.~Ideker, P.~Mali, L.~S. Qi \emph{et~al.}, ``Genetic interaction mapping in
  mammalian cells using crispr interference,'' \emph{Nature Methods}, 2017.

\bibitem{han2017synergistic}
K.~Han, E.~E. Jeng, G.~T. Hess, D.~W. Morgens, A.~Li, and M.~C. Bassik,
  ``Synergistic drug combinations for cancer identified in a crispr screen for
  pairwise genetic interactions,'' \emph{Nature Biotechnology}, 2017.

\bibitem{boucher2013genetic}
B.~Boucher and S.~Jenna, ``Genetic interaction networks: better understand to
  better predict,'' \emph{Frontiers in genetics}, vol.~4, 2013.

\bibitem{zhan2016towards}
T.~Zhan and M.~Boutros, ``Towards a compendium of essential genes--from model
  organisms to synthetic lethality in cancer cells,'' \emph{Critical reviews in
  biochemistry and molecular biology}, vol.~51, no.~2, pp. 74--85, 2016.

\bibitem{deutscher2008can}
D.~Deutscher, I.~Meilijson, S.~Schuster, and E.~Ruppin, ``Can single knockouts
  accurately single out gene functions?'' \emph{BMC Systems Biology}, vol.~2,
  no.~1, p.~50, 2008.

\bibitem{suthers2009genome}
P.~F. Suthers, A.~Zomorrodi, and C.~D. Maranas, ``Genome-scale gene/reaction
  essentiality and synthetic lethality analysis,'' \emph{Molecular systems
  biology}, vol.~5, no.~1, p. 301, 2009.

\bibitem{folger2011predicting}
O.~Folger, L.~Jerby, C.~Frezza, E.~Gottlieb, E.~Ruppin, and T.~Shlomi,
  ``Predicting selective drug targets in cancer through metabolic networks,''
  \emph{Molecular systems biology}, vol.~7, no.~1, p. 501, 2011.

\bibitem{pratapa2015fast}
A.~Pratapa, S.~Balachandran, and K.~Raman, ``Fast-sl: an efficient algorithm to
  identify synthetic lethal sets in metabolic networks,''
  \emph{Bioinformatics}, p. btv352, 2015.

\bibitem{jerby2014predicting}
L.~Jerby-Arnon, N.~Pfetzer, Y.~Y. Waldman, L.~McGarry, D.~James, E.~Shanks,
  B.~Seashore-Ludlow, A.~Weinstock, T.~Geiger, P.~A. Clemons \emph{et~al.},
  ``Predicting cancer-specific vulnerability via data-driven detection of
  synthetic lethality,'' \emph{Cell}, vol. 158, no.~5, pp. 1199--1209, 2014.

\bibitem{srihari2015inferring}
S.~Srihari, J.~Singla, L.~Wong, and M.~A. Ragan, ``Inferring synthetic lethal
  interactions from mutual exclusivity of genetic events in cancer,''
  \emph{Biology direct}, vol.~10, no.~1, p.~57, 2015.

\bibitem{sinha2017systematic}
S.~Sinha, D.~Thomas, S.~Chan, Y.~Gao, D.~Brunen, D.~Torabi, A.~Reinisch,
  D.~Hernandez, A.~Chan, E.~B. Rankin \emph{et~al.}, ``Systematic discovery of
  mutation-specific synthetic lethals by mining pan-cancer human primary tumor
  data,'' \emph{Nature Communications}, vol.~8, 2017.

\bibitem{kranthi2013identification}
T.~Kranthi, S.~Rao, and P.~Manimaran, ``Identification of synthetic lethal
  pairs in biological systems through network information centrality,''
  \emph{Molecular BioSystems}, vol.~9, no.~8, pp. 2163--2167, 2013.

\bibitem{zhang2015predicting}
F.~Zhang, M.~Wu, X.-J. Li, X.-L. Li, C.~K. Kwoh, and J.~Zheng, ``Predicting
  essential genes and synthetic lethality via influence propagation in
  signaling pathways of cancer cell fates,'' \emph{Journal of bioinformatics
  and computational biology}, vol.~13, no.~03, p. 1541002, 2015.

\bibitem{jacunski2015connectivity}
A.~Jacunski, S.~J. Dixon, and N.~P. Tatonetti, ``Connectivity homology enables
  inter-species network models of synthetic lethality,'' \emph{PLoS Comput
  Biol}, vol.~11, no.~10, p. e1004506, 2015.

\bibitem{wong2004combining}
S.~L. Wong, L.~V. Zhang, A.~H. Tong, Z.~Li, D.~S. Goldberg, O.~D. King,
  G.~Lesage, M.~Vidal, B.~Andrews, H.~Bussey \emph{et~al.}, ``Combining
  biological networks to predict genetic interactions,'' \emph{Proceedings of
  the National Academy of Sciences of the United States of America}, vol. 101,
  no.~44, pp. 15\,682--15\,687, 2004.

\bibitem{li2011understanding}
B.~Li, W.~Cao, J.~Zhou, and F.~Luo, ``Understanding and predicting synthetic
  lethal genetic interactions in saccharomyces cerevisiae using domain genetic
  interactions,'' \emph{BMC systems biology}, vol.~5, no.~1, p.~73, 2011.

\bibitem{pandey2010integrative}
G.~Pandey, B.~Zhang, A.~N. Chang, C.~L. Myers, J.~Zhu, V.~Kumar, and E.~E.
  Schadt, ``An integrative multi-network and multi-classifier approach to
  predict genetic interactions,'' \emph{PLoS Comput Biol}, vol.~6, no.~9, p.
  e1000928, 2010.

\bibitem{wu2014silico}
M.~Wu, X.~Li, F.~Zhang, X.~Li, C.-K. Kwoh, and J.~Zheng, ``In silico prediction
  of synthetic lethality by meta-analysis of genetic interactions, functions,
  and pathways in yeast and human cancer,'' \emph{Cancer informatics}, vol.
  Suppl. 3, p.~71, 2014.

\bibitem{deshpande2013comparative}
R.~Deshpande, M.~K. Asiedu, M.~Klebig, S.~Sutor, E.~Kuzmin, J.~Nelson,
  J.~Piotrowski, S.~H. Shin, M.~Yoshida, M.~Costanzo \emph{et~al.}, ``A
  comparative genomic approach for identifying synthetic lethal interactions in
  human cancer,'' \emph{Cancer research}, vol.~73, no.~20, pp. 6128--6136,
  2013.

\bibitem{guo2016synlethdb}
J.~Guo, H.~Liu, and J.~Zheng, ``Synlethdb: synthetic lethality database toward
  discovery of selective and sensitive anticancer drug targets,'' \emph{Nucleic
  acids research}, vol.~44, no.~D1, pp. D1011--D1017, 2016.

\bibitem{wang2013predicting}
H.~Wang, H.~Huang, C.~Ding, and F.~Nie, ``Predicting protein--protein
  interactions from multimodal biological data sources via nonnegative matrix
  tri-factorization,'' \emph{Journal of Computational Biology}, vol.~20, no.~4,
  pp. 344--358, 2013.

\bibitem{liu2016neighborhood}
Y.~Liu, M.~Wu, C.~Miao, P.~Zhao, and X.-L. Li, ``Neighborhood regularized
  logistic matrix factorization for drug-target interaction prediction,''
  \emph{PLoS Comput Biol}, vol.~12, no.~2, p. e1004760, 2016.

\bibitem{wang2017improved}
L.~Wang, X.~Li, L.~Zhang, and Q.~Gao, ``Improved anticancer drug response
  prediction in cell lines using matrix factorization with similarity
  regularization,'' \emph{BMC cancer}, vol.~17, no.~1, p. 513, 2017.

\bibitem{johnsonlogistic}
C.~C. Johnson, ``Logistic matrix factorization for implicit feedback data,''
  \emph{NIPS 2014 Workshop on Distributed Machine Learning and Matrix
  Computations}, 2014.

\bibitem{wang2017robust}
Y.~Wang, A.~Kucukelbir, and D.~M. Blei, ``Robust probabilistic modeling with
  bayesian data reweighting,'' in \emph{International Conference on Machine
  Learning}, 2017, pp. 3646--3655.

\bibitem{duchi2011adaptive}
J.~Duchi, E.~Hazan, and Y.~Singer, ``Adaptive subgradient methods for online
  learning and stochastic optimization,'' \emph{J Mach Learn Res}, vol.~12, pp.
  2121--2159, 2011.

\bibitem{liu2017learning}
Y.~Liu, P.~Zhao, X.~Liu, M.~Wu, L.~Duan, and X.-L. Li, ``Learning user
  dependencies for recommendation,'' in \emph{Proceedings of the 26th
  International Joint Conference on Artificial Intelligence}, 2017, pp.
  2379--2385.

\bibitem{gupta1999matrix}
A.~K. Gupta and D.~K. Nagar, \emph{Matrix variate distributions}.\hskip 1em
  plus 0.5em minus 0.4em\relax CRC Press, 1999, vol. 104.

\bibitem{li2014syn}
X.-j. Li, S.~K. Mishra, M.~Wu, F.~Zhang, and J.~Zheng, ``Syn-lethality: an
  integrative knowledge base of synthetic lethality towards discovery of
  selective anticancer therapies,'' \emph{BioMed research international}, vol.
  2014, 2014.

\bibitem{schmidt2013genomernai}
E.~E. Schmidt, O.~Pelz, S.~Buhlmann, G.~Kerr, T.~Horn, and M.~Boutros,
  ``Genomernai: a database for cell-based and in vivo rnai phenotypes, 2013
  update,'' \emph{Nucleic acids research}, vol.~41, no.~D1, pp. D1021--D1026,
  2013.

\bibitem{keshava2009human}
T.~Keshava~Prasad, R.~Goel, K.~Kandasamy, S.~Keerthikumar, S.~Kumar,
  S.~Mathivanan, D.~Telikicherla, R.~Raju, B.~Shafreen, A.~Venugopal
  \emph{et~al.}, ``Human protein reference database—2009 update,''
  \emph{Nucleic acids research}, vol.~37, no. suppl\_1, pp. D767--D772, 2009.

\bibitem{chua2006exploiting}
H.~N. Chua, W.-K. Sung, and L.~Wong, ``Exploiting indirect neighbours and
  topological weight to predict protein function from protein--protein
  interactions,'' \emph{Bioinformatics}, vol.~22, no.~13, pp. 1623--1630, 2006.

\bibitem{wang2007new}
J.~Z. Wang, Z.~Du, R.~Payattakool, P.~S. Yu, and C.-F. Chen, ``A new method to
  measure the semantic similarity of go terms,'' \emph{Bioinformatics},
  vol.~23, no.~10, pp. 1274--1281, 2007.

\bibitem{mei2012drug}
J.-P. Mei, C.-K. Kwoh, P.~Yang, X.-L. Li, and J.~Zheng, ``Drug--target
  interaction prediction by learning from local information and neighbors,''
  \emph{Bioinformatics}, vol.~29, no.~2, pp. 238--245, 2012.

\bibitem{davis2006relationship}
J.~Davis and M.~Goadrich, ``The relationship between precision-recall and roc
  curves,'' in \emph{Proceedings of the 23rd international conference on
  Machine learning}.\hskip 1em plus 0.5em minus 0.4em\relax ACM, 2006, pp.
  233--240.

\bibitem{chen2016xgboost}
T.~Chen and C.~Guestrin, ``Xgboost: A scalable tree boosting system,'' in
  \emph{Proceedings of the 22Nd ACM SIGKDD International Conference on
  Knowledge Discovery and Data Mining}.\hskip 1em plus 0.5em minus 0.4em\relax
  ACM, 2016, pp. 785--794.

\end{thebibliography}


\begin{IEEEbiography}[{\includegraphics[width=1in,height=1.25in,clip,keepaspectratio]{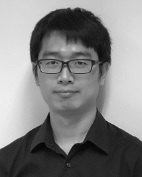}}]{Yong Liu} is currently a Research Scientist in the Joint NTU-UBC Research Centre of Excellence in Active Living for the Elderly (LILY), Nanyang Technological University, Singapore. He received his B.S. from University of Science and Technology of China in 2008 and Ph.D. from Nanyang Technological University in 2016. He was a Data Scientist in NTUC Enterprise, Singapore from November 2017 to July 2018, and a Research Scientist in the Data Analytics Department at the Institute for Infocomm Research (I2R), A*STAR, Singapore from November 2015 to October 2017. His research areas include recommender systems, social media mining, and Bioinformatics. His research papers appear in leading international conferences and journals. He has been invited as a PC member of major conferences such as KDD, IJCAI, AAAI, CIKM, ICDM, and reviewer for IEEE/ACM transactions. He is a member of ACM.
\end{IEEEbiography}

\begin{IEEEbiography}[{\includegraphics[width=1in,height=1.25in,clip,keepaspectratio]{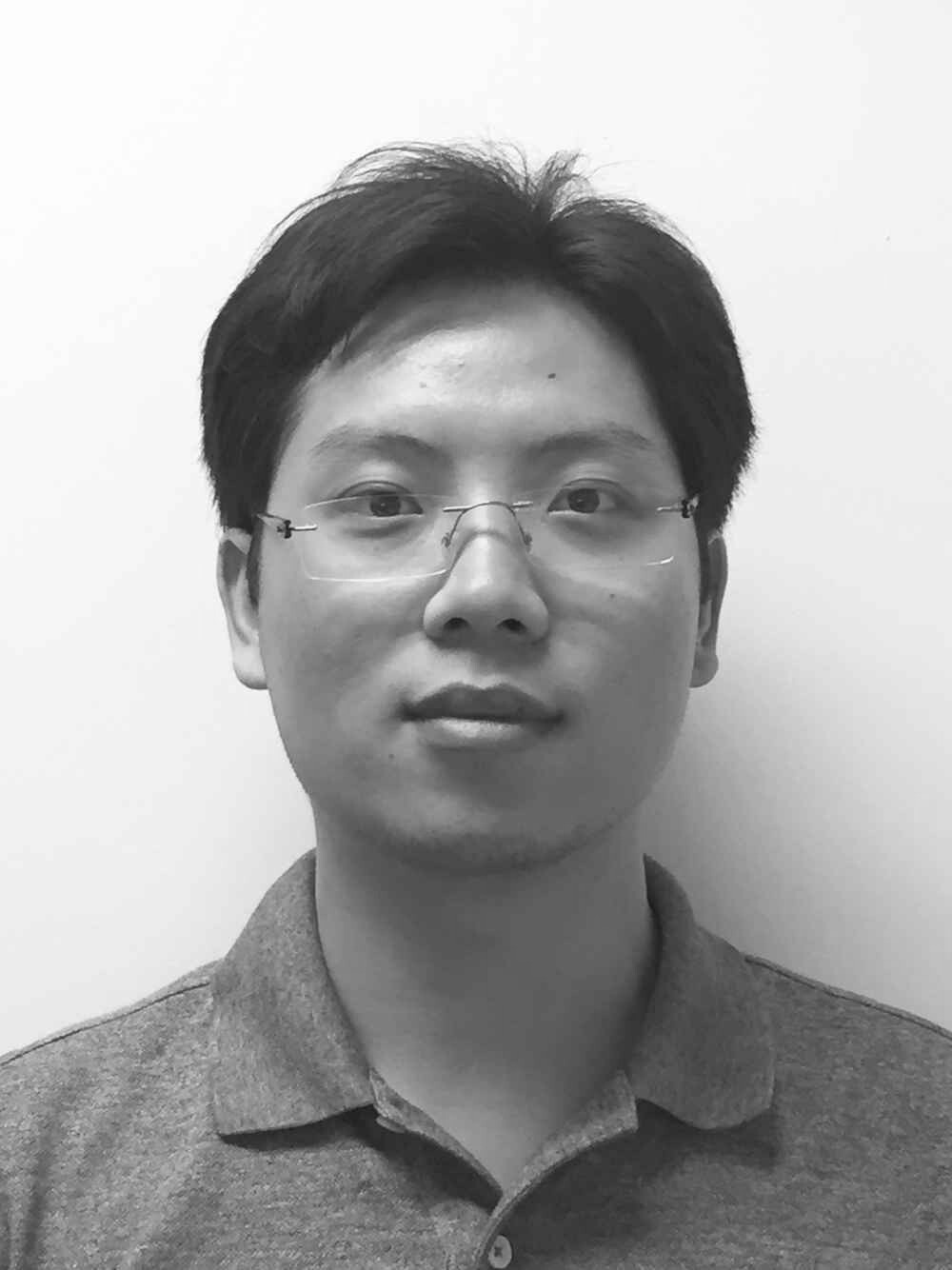}}]{Min Wu} is currently a Research Scientist in the Data Analytics Department at the Institute for Infocomm Research (I2R) under the Agency for Science, Technology and Research (A*STAR), Singapore. He received the B.Eng. from the University of Science and Technology of China (USTC), China in 2006 and his Ph.D. degree from Nanyang Technological University, Singapore in 2011. He received the best paper awards in the 15th International Conference on Bioinformatics (InCoB 2016) and the 20th International Conference on Database Systems for Advanced Applications (DASFAA 2015). He also won the IJCAI contest 2015 on repeated buyers prediction after sales promotion. His current research interests include machine learning, data mining and bioinformatics.
\end{IEEEbiography}

\begin{IEEEbiography}[{\includegraphics[width=1in,height=1.25in,clip,keepaspectratio]{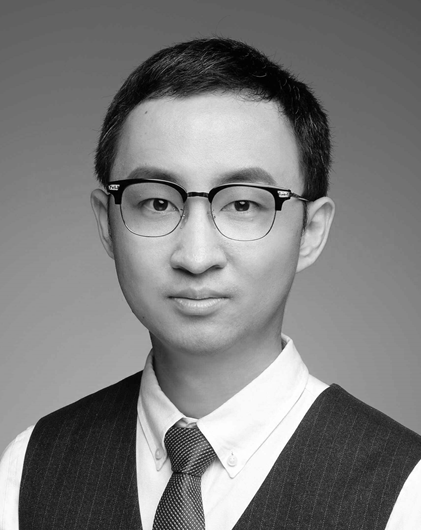}}]{Chenghao Liu} is a Postdoctoral Research Fellow in the School of Information Systems (SIS), Singapore Management University (SMU), Singapore.  He received his Bachelor degree and Ph.D degrees from the Zhejiang Uniersity. His research interests include large-scale machine learning (online learning and deep learning) with application to tackle big data analytics challenges across a wide range of real-world applications.
\end{IEEEbiography}

\begin{IEEEbiography}[{\includegraphics[width=1in,height=1.25in,clip,keepaspectratio]{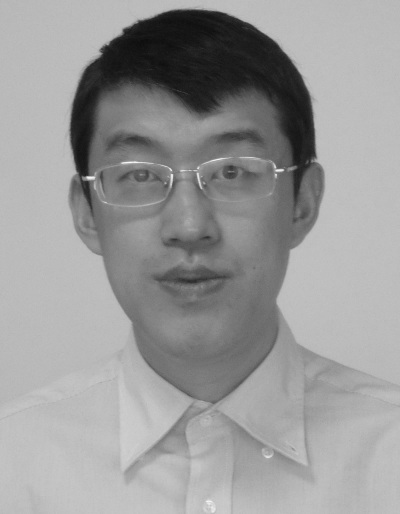}}]{Xiaoli Li} is currently head of Data Analytics Department at the Institute for Infocomm Research, A*STAR, Singapore. He also holds adjunct associate professor positions at the National University of Singapore and Nanyang Technological University. His research interests include data mining, machine learning and bioinformatics. He has served as a PC member/workshop chair/session chair in leading data mining related conferences (including KDD, ICDM, SDM, PKDD/ECML, PAKDD, WWW, AAAI, and CIKM) and as an editor of bioinformatics-related books. In 2005, he received the Best Paper Award in the 16th International Conference on Genome Informatics (GIW 2005). In 2011, he received the Best Paper Runner-Up Award in the 16th International Conference on Database Systems for Advanced Applications (DASFAA 2011). Xiaoli has published more than 170 papers, including top tier conferences such as KDD, ICDM, SDM, PKDD/ECML, ICML, IJCAI, AAAI, ACL, EMNLP, SIGIR, CIKM, UbiCom, etc. as well as some top tier journals such as IEEE Transactions TKDE, Bioinformatics and IEEE Transactions on Reliability.
\end{IEEEbiography}

\begin{IEEEbiography}[{\includegraphics[width=1in,height=1.25in,clip,keepaspectratio]{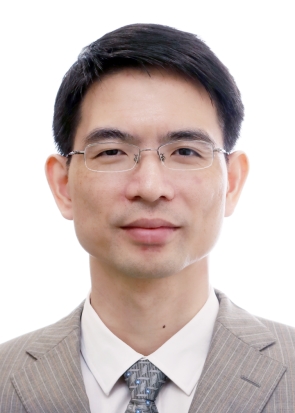}}]{Jie Zheng} is an Associate Professor at the School of Information Science and Technology, ShanghaiTech University, Shanghai, China. He received his B. Eng (honors) in 2000 from Zhejiang University in China, and his Ph.D. in 2006 from the University of California, Riverside in USA, both in Computer Science. From 2006 to 2011, he was a Postdoctoral Visiting Fellow and Research Associate at the National Center for Biotechnology Information (NCBI), National Library of Medicine (NLM), National Institutes of Health (NIH), USA. From Feb. 2011 to July 2018, he was an Assistant Professor at the School of Computer Science and Engineering, Nanyang Technological University (NTU), Singapore. Dr. Zheng's research interests include bioinformatics, computational genomics and systems biology, and biomedical data science. He develops novel computational methods (e.g. machine learning and data mining algorithms, artificial intelligence techniques, dynamical and data-driven models) to help answer biomedical questions. While trained as a Computer Scientist, Dr. Zheng maintains active and long-standing collaborations with Life Scientists.
\end{IEEEbiography}

\end{document}